\documentclass[letterpaper, 10 pt, conference]{ieeeconf}  %

\IEEEoverridecommandlockouts                              %

\usepackage{float}
\usepackage{graphics}
\usepackage{amsmath}
\usepackage{amssymb}  
\usepackage{graphicx}
\usepackage{tablefootnote}
\usepackage{multirow}
\usepackage[table,xcdraw]{xcolor}
\usepackage{bm}
\usepackage{booktabs}
\usepackage{lineno}
\usepackage{svg}

\title{\LARGE \bf
U-BEV: Height-aware Bird's-Eye-View Segmentation and Neural Map-based Relocalization}

\author{Andrea Boscolo Camiletto$^{1,2}$, Alfredo Bochicchio$^{3}$, Alexander Liniger$^{3}$, Dengxin Dai$^{3}$, Abel Gawel$^{3}$%
\thanks{$^{1}$Work done while at BayerLab, Huawei Zurich Research Center.}%
\thanks{$^{2}$MPI for Informatics. {\tt\small aboscolo@mpi-inf.mpg.de}}%
\thanks{$^{3}$BayerLab, Huawei. {\tt\small firstname.lastname@huawei.com}}%
}

\usepackage[backend=biber,style=ieee]{biblatex} %
\usepackage{caption}
\begin{document}

\maketitle
\thispagestyle{empty}
\pagestyle{empty}

\begin{abstract}
Efficient relocalization is essential for intelligent vehicles when GPS reception is insufficient or sensor-based localization fails. Recent advances in Bird's-Eye-View (BEV) segmentation allow for accurate estimation of local scene appearance and in turn, can benefit the relocalization of the vehicle. However, one downside of BEV methods is the heavy computation required to leverage the geometric constraints. This paper presents U-BEV, a U-Net inspired architecture that extends the current state-of-the-art by allowing the BEV to reason about the scene on multiple height layers before flattening the BEV features. We show that this extension boosts the performance of the U-BEV by up to $4.11$ IoU. Additionally, we combine the encoded neural BEV with a differentiable template matcher to perform relocalization on neural SD-map data. The model is fully end-to-end trainable and outperforms transformer-based BEV methods of similar computational complexity by $1.7$ to $2.8$ mIoU and BEV-based relocalization by over $26\%$ Recall Accuracy on the nuScenes dataset.

\end{abstract}

\section{INTRODUCTION}
\label{sec:intro}

Image-based relocalization and place recognition are among the fundamental challenges for autonomous agents and have been active research topics in the robotics community for over a decade~\cite{lowry2015visual, garg2021your}. 
Efficient and accurate relocalization is especially relevant in Autonomous Driving, where agents travel tens of meters per second in constantly changing environments. The need for relocalization arises when GPS signals are unreliable, such as in densely populated cities where Urban Canyons are common. While the relocalization problem is well studied with high-fidelity map data and onboard sensing, the need for using camera-only systems and lightweight map representations drives contemporary research. Additionally, algorithms need fast inference times with strong generalization capabilities and cover degenerate localization cases where only weak perceptual cues are present. Current image-based methods however are either accurate but prohibitively expensive~\cite{m2bev, bevfusion, zhang2023occformer,wei2023surroundocc}, suffer under degenerate cases~\cite{zhou2022visual}, or rely on highly curated~\cite{zhang2022bev} or heavy representations~\cite{zhu2023nemo}. 

Recent advances in Bird's-Eye-View (BEV)-based scene understanding boosted the perception capabilities of autonomous agents~\cite{bevfusion, cvt, lss, monosemantic, bevstitch, pon}. These representations lend themselves well to localization cases as they are trained on common autonomous driving map data.
Hence, new BEV-based localization methods~\cite{zhang2022bev, orienternet, sarlin2023snap} elegantly extend cross-view localization by predicting from camera images a local BEV-representation on the 2D ground plane and matching it to a semantic map. However, the flat world assumption reduces BEV accuracy in many 3D real-world cases and lowers localization accuracy.

\begin{figure}
\centering
\includegraphics[width=0.48\textwidth]{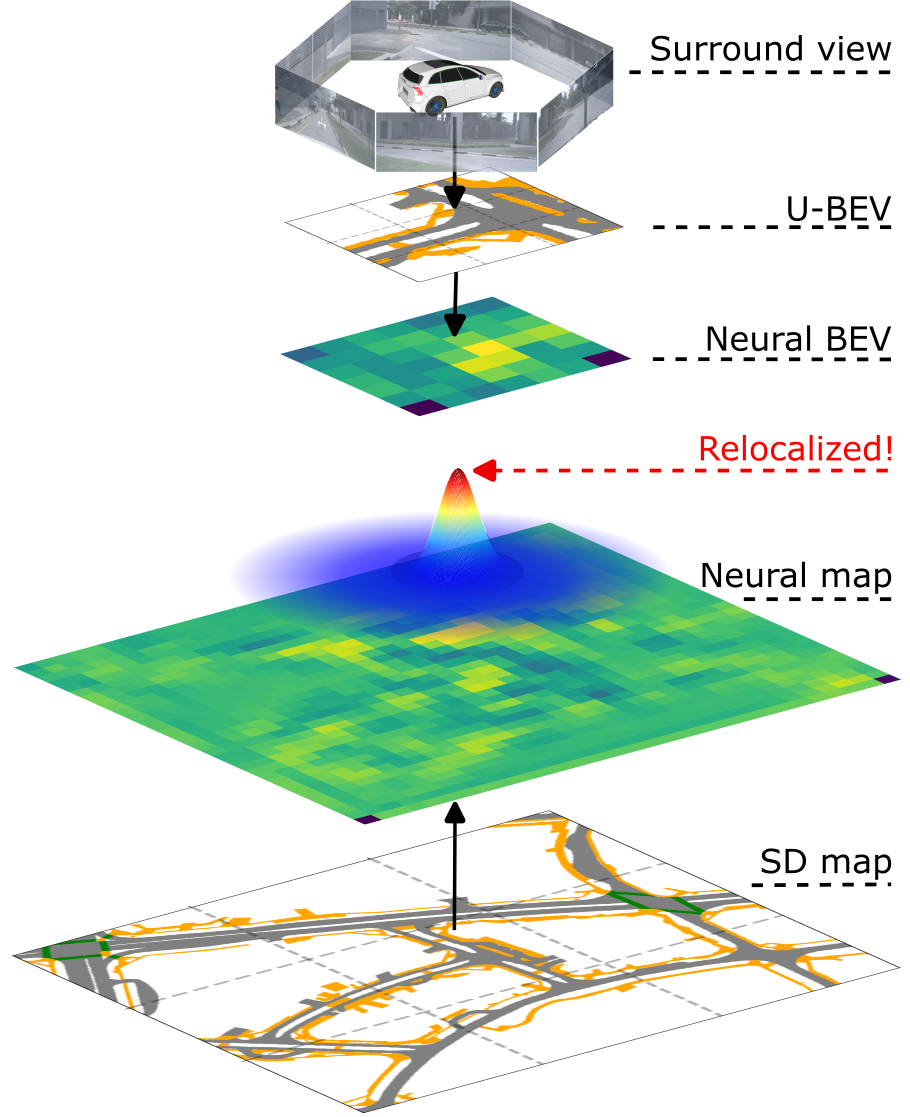}
\caption{U-BEV proposes a novel BEV representation from surround-view images for efficient neural relocalization in SD map data.}
\label{fig:teaser}
\end{figure}

In this work, we propose a new approach for estimating BEVs that leverages height-aware feature embedding, allowing the network to reason on the depth dimension without the need for heavy computations. The base architecture leverages an encoder-decoder structure with an additional geometrical projection layer to reason both in the image (during the encoding) and BEV space (during the decoding). We advocate for a  2-step localization approach, where autonomous agents first globally estimate their location within several meters and then rely on local methods to obtain application-required accuracies. Therefore, we utilize lightweight Standard Definition (SD) map data and aim for accuracies of one-shot relocalization below $10m$. In this work, we integrate our BEV representation with a differentiable template matcher, which allows for end-to-end training and real-time relocalization. The localization architecture can work on arbitrary BEV methods by encoding them and the corresponding map data into a Neural representation. This work, combining U-BEV with the relocalization module outperforms both other BEV methods and contemporary BEV-based localization on the nuScenes dataset by over $26.4\%$ Recall Accuracy at $10m$. In summary, this paper presents the following contributions:
\begin{enumerate}
    \item A new lightweight U-BEV architecture that is geometrically constrained and leverages the height of the points from the ground instead of their depth from the camera.
    \item An end-to-end trainable real-time global localization algorithm between Neural BEV and Neural encoded SD maps.
    \item Improved BEV ($1.7$ to $2.8$ higher IoU) and localization performance ($+26.4\%$ Recall Accuracy at $10m$) on the nuScenes dataset.
\end{enumerate}

\section{RELATED WORK}
This publication contributes to the state-of-the-art in BEV estimation and image-based relocalization. We review related work in the following.
\subsection{Bird's-Eye-View Segmentation}
Semantic Segmentation of the car's surroundings in BEV is a common task in the autonomous driving domain.
The goal is to flatten and discretize the environment around the car into tiles of fixed size and segment each tile into labels e.g., road, sidewalk, building, etc.

This segmentation is often addressed as a side task to the goal of 3D object detection and leverages both camera data and LiDAR input ~\cite{bevfusion, pointpainting, mvp}. Recently, the community is shifting towards addressing the problem with cameras only~\cite{monosemantic, m2bev, cvt, crossviewss, fiery, sta}.

In this shift, handling depth is the main differentiator. To reason in BEV space, pixels require projection from the image to the world coordinates. This in turn requires spatial understanding.
One line of works leverages a discretized explicit depth estimation~\cite{lss, bevfusion, m2bev}, which is used to weight the features along the ray accordingly. Another option is to learn a mapping from image space to BEV space via a transformer or attention layer like~\cite{cvt, panobev}. Finally, features can also be warped to the BEV space using homographies explicitly~\cite{bevstitch} or learned transformations~\cite{pon}.

While predicting the depth and projecting accordingly is undeniably a desirable geometric constraint, it can be resource-intensive. A common trend is to discretize the depth in the range of $[1, 60]$ meters with a step size of $0.5$ meter~\cite{bevfusion, bevdet} which leads to the creation of a very dense point cloud representation. %
In this work, we propose a different direction: instead of predicting dense depth, we estimate a sparse set of height layers.

\subsection{Image-based Relocalization}
Image-based relocalization is closely related to Visual Place Recognition which defines the problem of retrieving a query image from a database~\cite{garg2021your, arandjelovic2016netvlad} and estimating the (relative) pose. The most popular approach is to estimate local feature embeddings per image and either query them in a database (map)~\cite{arandjelovic2016netvlad} or regress the map localization by spatially rolling out image features. When viewpoints are vastly different as is the case in this work, the problem is coined as cross-view localization.
Recent works bridge the viewpoint difference using a sparse spatial-semantic representation~\cite{wang2022ltsr,gawel2018x,zhou2022visual}, leverage generative networks that generate ground-view images from aerial views and perform database retrieval~\cite{toker2021coming, lu2020geometry}, or learn a joint feature embedding for different views~\cite{fervers2023uncertainty, wang2023satellite, xia2022visual, lentsch2023slicematch, zhao2023co, lin2022joint, yang2021cross}. 
Another line of work lifts this challenge to the semantic domain to further bridge the perceptual gap between ground-view imagery (e.g., surround-view images from a car) to top-down map data. They generate BEV representations from ground-view images and match them to a learned or off-the-shelf semantic map~\cite{zhang2022bev, zhou2022visual, orienternet, sarlin2023snap}. This work follows a similar approach, but in contrast to~\cite{orienternet, sarlin2023snap} employs the more powerful U-BEV representation that allows localization in lower detailed maps. The proposed work also bridges large-scale localization errors of hundreds of meters, larger than~\cite{zhang2022bev} which addresses $2m$ errors, and covers more degenerate cases than~\cite{zhou2022visual}, e.g., our method is able to localize on open roads without distinct landmarks such as poles or signs.

\section{METHOD}
\begin{figure*}[!htb]
\centering
\includegraphics[width=0.95\textwidth]{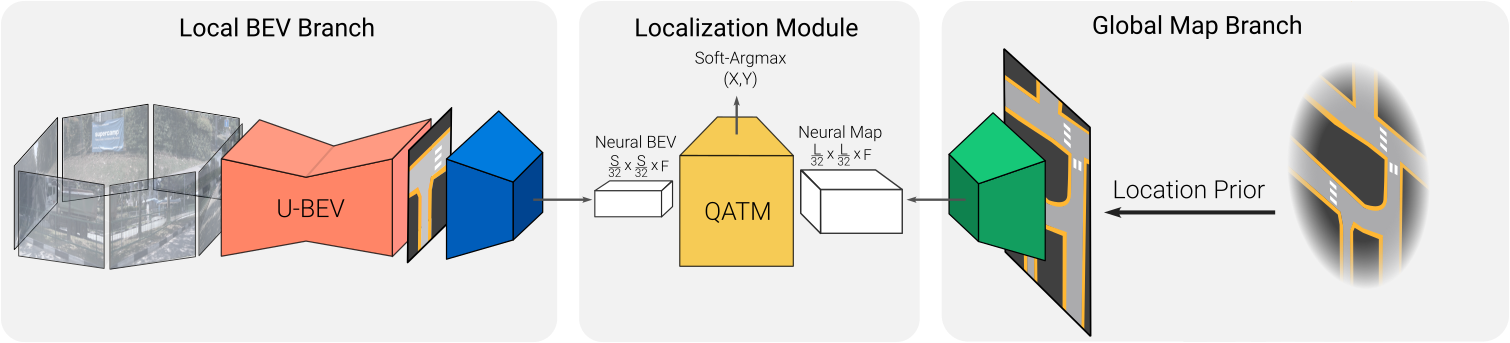}
\caption{Overview of the U-BEV Neural Relocalization Model. U-BEV predicts the local BEV from a set of surround cameras. A pretrained encoder extract features from it yielding the Neural BEV (left). The map encoder extracts features from the cropped global SD map based on Location prior $\xi_{init}$ (right) to build the neural map representation. The deep template matching module (QATM) computes the best matching location (center).}
\label{fig:overview}
\end{figure*}

The proposed full algorithm localizes a set of surround-view images in an SD map. It generates a local BEV representation from the surround images and a neural map encoding from an SD map tile given a coarse 3D location prior $\xi_{init}=(x_{init}, y_{init}, \phi_{init})$ from onboard sensors (e.g. a noisy GPS signal and a compass). A deep template matcher then slides the local Neural BEV over the global Neural map resulting in a similarity map. The localization finishes by returning the Soft-Argmax of the similarity map. An overview of our method is illustrated in Fig.~\ref{fig:overview}.

\subsection{Bird's Eye View Reconstruction}
\begin{figure*}[!htb]
\centering
\includegraphics[width=0.95\textwidth]{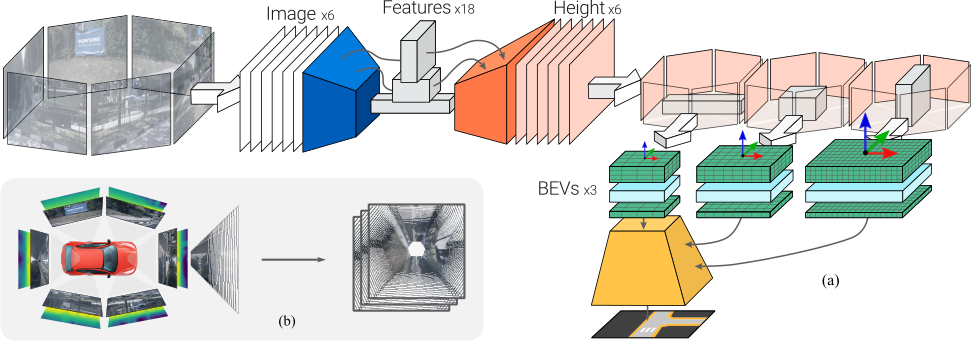}
\caption{Architecture of the U-BEV Model. (a) The pretrained backbone (in blue) extracts features from all 6 cameras around the car. The first decoder (in orange) predicts the height of each pixel on each input image. This height is used to project features from different cameras into a single BEV (in green). Deeper features get projected to lower-resolution BEVs and are then upsampled in an encoder-decoder fashion with skip connections (in yellow). (b) Illustrates the projection operation from surround view images and heights to different BEV layers. }
\label{fig:ubev}
\end{figure*}

We propose a novel lightweight and accurate BEV architecture to reconstruct the car's surroundings from a set of surround-view images.
Our model, called U-BEV, is structured as an Encoder-Decoder with an in-between layer that geometrically projects the extracted features of the bottleneck and skip connections from the image to the BEV space. An overview can be seen in Fig.~\ref{fig:ubev}.

Given a set of 6 images and their intrinsic and extrinsic calibrations, we predict a BEV $\mathbf B \in \mathbb{R}^{S\times S \times N}$, where S is the size of the BEV in pixel and N is the number of labels available in the map. We use the center of the back wheel axis as our origin point following the convention in the nuScenes dataset~\cite{nuscenes}.

\textbf{Feature Extraction:} We extract features at different resolutions from all 6 images with a light-weight pretrained EfficientNet~\cite{effnet} backbone, a common approach in smaller models ~\cite{lss, cvt}. Specifically, we extract features at strides $\times2, \times4, \times8, \times16$, intentionally omitting the last stride for computational reasons. The extracted features are used as skip connections throughout the architecture. (Blue box in Fig.~\ref{fig:ubev})

\begin{figure}[!htb]
\centering
\includegraphics[width=0.49\textwidth]{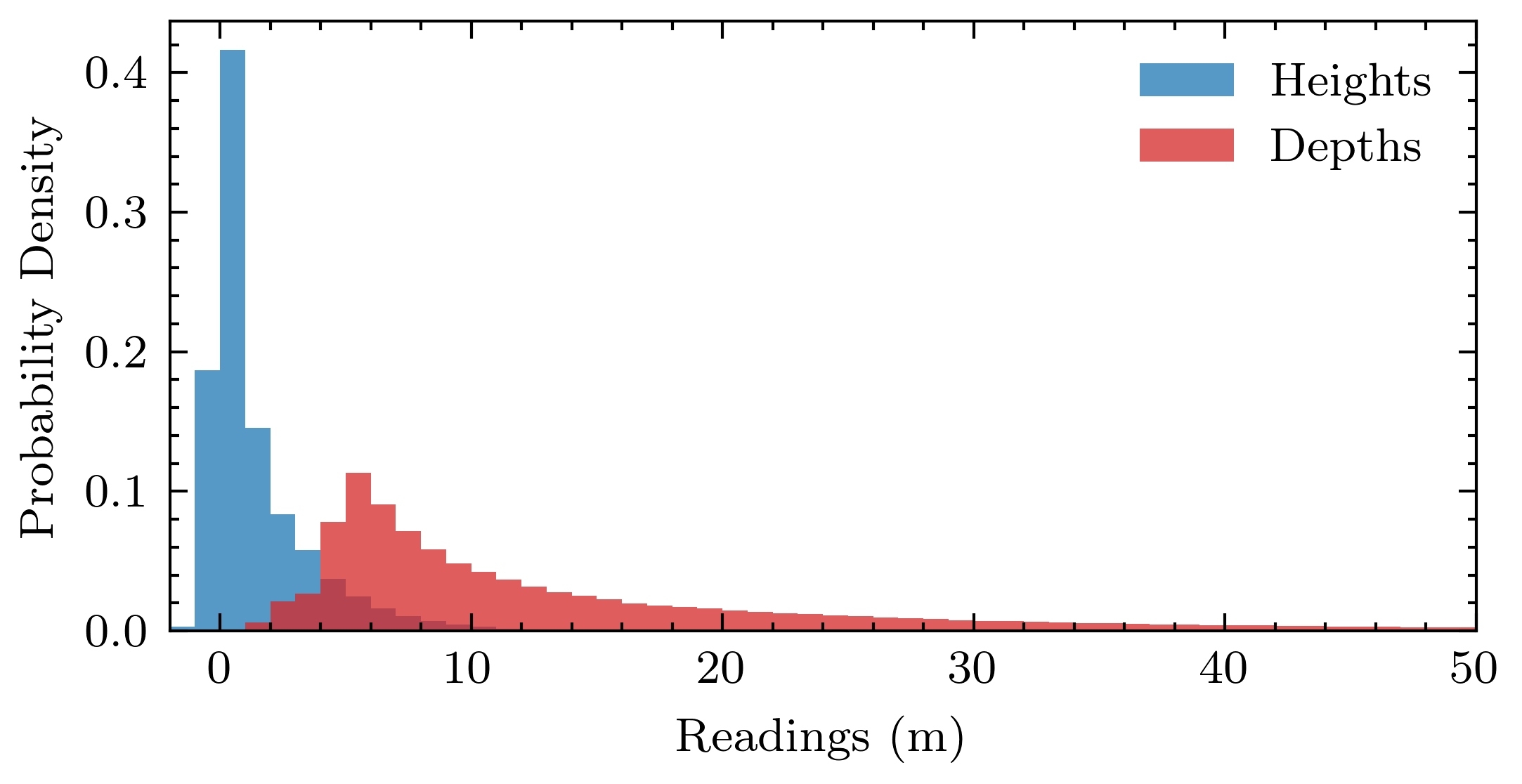}
\caption{Distribution of LiDAR readings reprojected on image planes when written as the height from the ground of the car frame and as the distance from the camera, on nuScenes.}
\label{fig:hvsd}
\end{figure}

\textbf{Height Prediction:} A key contribution of U-BEV is height from the ground estimation to reason in 3D space. We predict heights leveraging the extracted features and a lightweight decoder~\cite{pan} to perform this pixel-wise operation (Orange part of Fig.~\ref{fig:ubev}). Contrary to a widespread practice in the BEV literature to predict implicitly or explicitly the depth we argue that the height of each observed pixel is a more efficient representation. This is motivated by the observation that high resolution in the $x,y$ ground plane is required for driving applications, while the vertical axis can be more coarsely discretized. Moreover, as shown in Fig.~\ref{fig:hvsd}, the depth is usually distributed over a longer range, e.g., $[0\text{-}50]$m, which requires a large number of discretization intervals. Height can be meaningfully discretized in a lower range, e.g., $[0\text{-}5]$m to interpret the surroundings. A lower number of bins has a direct impact throughout the model: it significantly reduces the complexity of the projection which can be slow and inefficient as shown by~\cite{bevfusion} (in our case up to $\times 20$) and also lowers the memory footprint.
Finally, the most relevant information, i.e. road surface, markings, curbs, etc. is packed in the lower part of that range.

We therefore set up the height prediction task as a classification problem using only $\mathbf{b} = [\text{-}0.5, 0, 0.5, 1, 2, 3]$ as bins. Modeling heights outside this range increase the computation complexity and does not provide significant improvements in our testing. More in details, our decoder outputs a prediction $\mathbf{H} \in \mathbb{R}^{H\times W\times 6}$ where $H,W$ is the shape of the input images.
The actual height prediction $\mathbf{\Psi}_h(i,j)$ of a certain pixel at index $(i,j)$ can be obtained through:
\begin{equation}\label{eq:height}
    \mathbf{\Psi}_h(i,j)=\sum_{k=1}^6\mathbf{b}_{k} \mathbf{H}(i,j,k)\,.
\end{equation}
We leverage this discretized height prediction to weigh each feature according to the likelihood of being in that bin. 

\textbf{Projection:} We project deeper - lower resolution - features in coarser BEV, and earlier - high resolution - features in higher resolution BEV. This allows us to upscale the BEV in the classic encoder-decoder fashion where the more detailed BEVs act as skip-connections (Green part in Fig.~\ref{fig:ubev}). 

We apply a modified Inverse Projective Mapping (IPM) to roll out the features from image coordinates to the BEV coordinates at different heights. (see Fig.~\ref{fig:ubev} b). To project each feature $\mathbf f \in \mathbb R^C$ from pixel $(u,v)$, we use the known extrinsic projection matrix $\mathbf P \in \mathbb R ^{3\times4}$ and the intrinsic parameters of the camera $\mathbf K \in \mathbb R ^{3\times3}$. To project at height $h$, we utilize the translation transformation in matrix form $\mathbf T_h \in \mathbb R ^{4\times4}$ to elevate the reference system to the desired height and perform the standard IPM at $z=0$.

The IPM formulation relates these variables to
\begin{align}
    \lambda
    \begin{bmatrix}
        u \\
        v \\
        1
    \end{bmatrix}
    = \mathbf K \cdot \mathbf P \cdot \mathbf T_h 
    \begin{bmatrix}
        x \\
        y \\
        0 \\
        1
    \end{bmatrix}\,.
\end{align}
This formulation conveniently allows dropping the third column of the $\mathbf K \cdot \mathbf P \cdot \mathbf T$ matrix, which allows us to invert it and solve for $x,y$. The operation can be parallelized on a GPU on all the features and applied and performed at all heights, thus resulting in an occupancy volume that can be weighted with the predicted distribution of each feature to be at a certain height.

\textbf{BEV Decoding:} Lastly, we squeeze the height dimension with two convolutional layers for each BEV. By keeping the ratio of the resolutions and channels constant, we can plug them into a classic decoder style with skip connections, yielding the final BEV output (Yellow part in Fig.~\ref{fig:ubev}).
 
\subsection{Map encoding}
Maps are passed into our system as rasterized boolean $N$-channels discretized surfaces, where $N$ is the number of classes, i.e. each semantic class is assigned a separate channel. In the case of polygon representations, as common in SD maps for autonomous driving, we preprocess the map by rasterizing the polygons of each class into channels.  %

\subsection{Localization}

To perform localization, we leverage both a local BEV $\mathbf{B}$ of the surroundings produced by the U-BEV model and a map tile of the global map $\mathbf{M}_{loc} \in \mathbb R ^{L\times L\times N}$ cropped given a coarse location prior $\xi_{init}$. Localization is achieved by differentiable template-matching in the latent space. To compensate imperfection of the local BEV reconstruction, the localization module extracts a neural representation from both the map tile and local BEV and builds a probability map $\mathbf M_{prob} \in \mathbb R ^{L\times L}$ over the map tile following~\cite{qatm}.

Matching the Neural BEV prediction and the Neural map in the feature space robustifies the localization module against errors and imperfections in the local BEV that may be due to occlusion, or perceptual degradation (e.g., lack of illumination, or bad weather) at the cost of resolution. To extract those features, we use the EfficientNet architecture as in the U-BEV model.

We apply a softmax with $\mathbf{\tilde{M}} = \text{softmax}_{2D}(\mathbf{M}_{prob})$ to use as our probability distribution. We extract a prediction by performing a soft-argmax \cite{softargmax} in both $x$ and $y$ direction with

\begin{equation}
    \Phi(\mathbf{\tilde{M}}) = \sum_{i,j}^S 
    (\mathbf{W} \otimes \mathbf{\tilde{M}})_{ij} \quad \text{with} \quad
    \begin{aligned}
    &\mathbf{W} \in \mathbb{R}^{S\times S \times 2} \\
    &\mathbf{W}_{ij}=(i,j)
    \end{aligned}\,.
\end{equation} 

\begin{figure*}[!ht]
\centering
\includegraphics[width=0.95\textwidth]{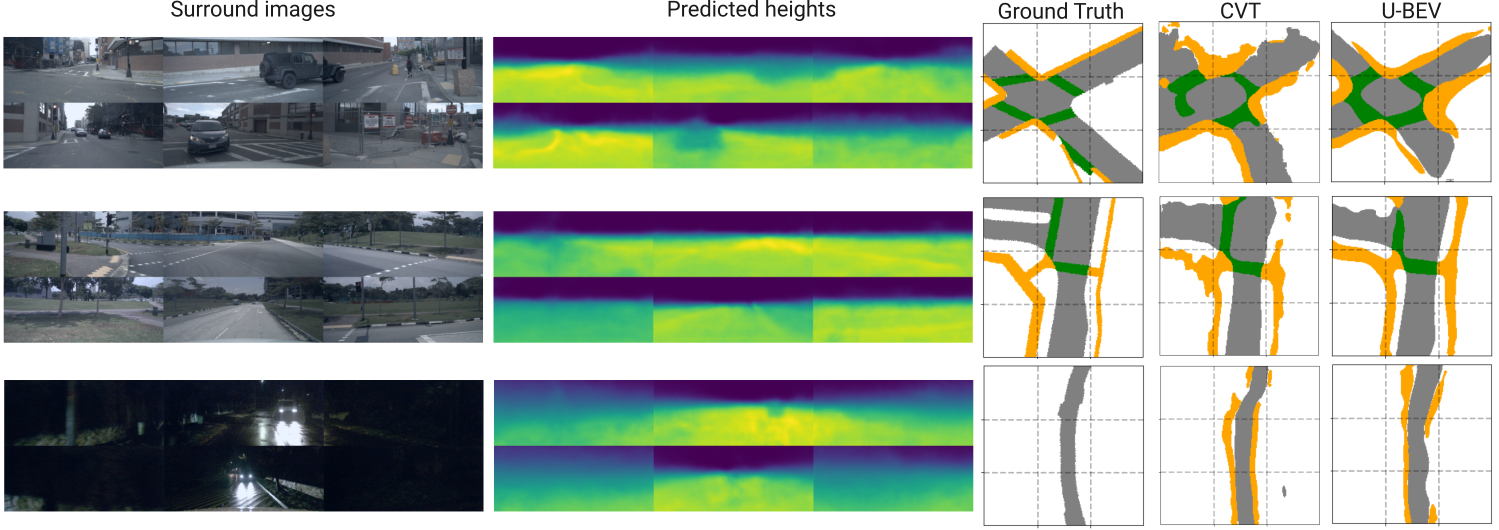}
\caption{Sample inputs and outputs of the proposed U-BEV, including surround images, predicted heights, and predicted and ground truth BEV. Compared to CVT, U-BEV more truthfully reconstructs drivable surfaces and sidewalks.}
\label{fig:ubev-samples}
\end{figure*}

\section{EXPERIMENTS}
We evaluate the proposed U-BEV algorithm individually and the full relocalization system including deep template matching on the public nuScenes driving dataset~\cite{nuscenes} with its map expansion. In our evaluation, we use surround camera data, ground truth poses, and ground truth SD map information. We use three main categories of the map data, i.e. drivable surface, crosswalks, and walkways. Since one goal is to investigate degenerate cases like feature-poor scenarios, we avoid the use of distinct objects, such as traffic lights in the localization scenario to force the network to learn about road network shape over distinct landmarks.

\subsection{Dataset}
The nuScenes dataset is the only publicly available one that provides 6-views surrounding images and SD maps aligned with the recorded data. It consists of 33989 data samples from 2 different cities.
The standard data split of nuScenes separates the sequences temporally (i.e. samples recorded next to each other are in the same split). This is well suited for most tasks that deal with dynamic objects as object detection, but suboptimal when static objects are involved as in the relocalization scenario. This creates data contamination: samples from different sequences on the same road are distributed into training and validation sets while observing the same part of the scene. We notice when training on this split that the models can learn unreasonably well and can predict parts of the maps that are not observable due to occlusion. Therefore, we use the split proposed by~\cite{roddick2020predicting}. This split separates samples both temporally and spatially and avoids overlaps in training and validation sets. We thus have 28008 data samples for training and 5981 for testing. 

\subsection{Experimental Setup}
\textbf{BEV:} For evaluating the U-BEV performance, we build BEV segmentation from 6 surround images of $100m\times 100m$ at a resolution of $50\text{cm}/\text{px}$. The sets of 6 images of resolution $1600\times 900$ are first cropped on the top by $27\%$ and downsampled to a resolution of $544\times 224$. We compare U-BEV with Cross-View-Transformer~(CVT)~\cite{cvt} a state-of-the-art BEV network which we retrained in our split for a baseline comparison under similar conditions as U-BEV. Retraining heavy BEV models with large GPU memory footprints is beyond the scope of our comparison as we aim to find a lightweight solution suitable for online relocalization.

\textbf{Relocalization:} For relocalization, we perturb the ground-truth location of each set of surround images uniformly within $[0m,100m]$ and crop tiles from the map data around the perturbed location of size $300m\times 300m$. Images and map tiles are given to the end-to-end localizer which estimates the position. We compare our localizer with the variant that includes CVT as BEV network and OrienterNet~\cite{orienternet}. For the latter, we use their pretrained model and finetune it on a set of different map resolutions and BEV ranges on the nuScenes training set. Furthermore, we disable the rotational degree of freedom in OrienterNet and exchange the OpenStreetMap tiles with nuScenes map tiles in training and validation. We evaluate relocalization with Recall Accuracy (RA) at distance thresholds $1,2,5,10\text{m}$

\subsection{Training}

Our training pipeline is divided into two steps. First, we pre-train the U-BEV module to predict the surrounding BEV. Only then, do we train the end-to-end model. All models are trained on a single NVIDIA $GV100$ GPU with 32GB of VRAM.

\textbf{BEV:} To have a fair comparison with CVT we train one model per class in our experiments, but we use a single multi-purpose model in the end-to-end pipeline.
We use a weighted focal loss function~\cite{focalloss} on the BEV prediction and a sparse Huber loss to supervise the height estimated using Equation~\ref{eq:height}.

For data augmentation, we rotate the camera poses around the car by a random amount. This does not affect the feature extraction and height prediction but helps the BEV decoder avoid overfitting by constantly changing the camera view directions. We train for 12 epochs, using 1-cycle learning rate scheduler~\cite{onecycle} with a maximum learning rate of $1 \cdot 10^{-3}$ and AdamW as optimizer~\cite{adamw}.

\textbf{Relocalization:} Before training the relocalization part, we freeze the U-BEV image encoder and its height decoder. This allows us to avoid overfitting to the limited nuScenes dataset. As before, we find benefits in augmenting the dataset by rotating both the local BEV (by modifying the extrinsic parameters) and the global map (by rotating the map) by random amounts during training until the last two epochs, where we disable this process.

We train this model similar to U-BEV, for 12 epochs and with the same LR and optimizer.

\begin{figure*}[!ht]
\centering
\includegraphics[width=0.95\textwidth]{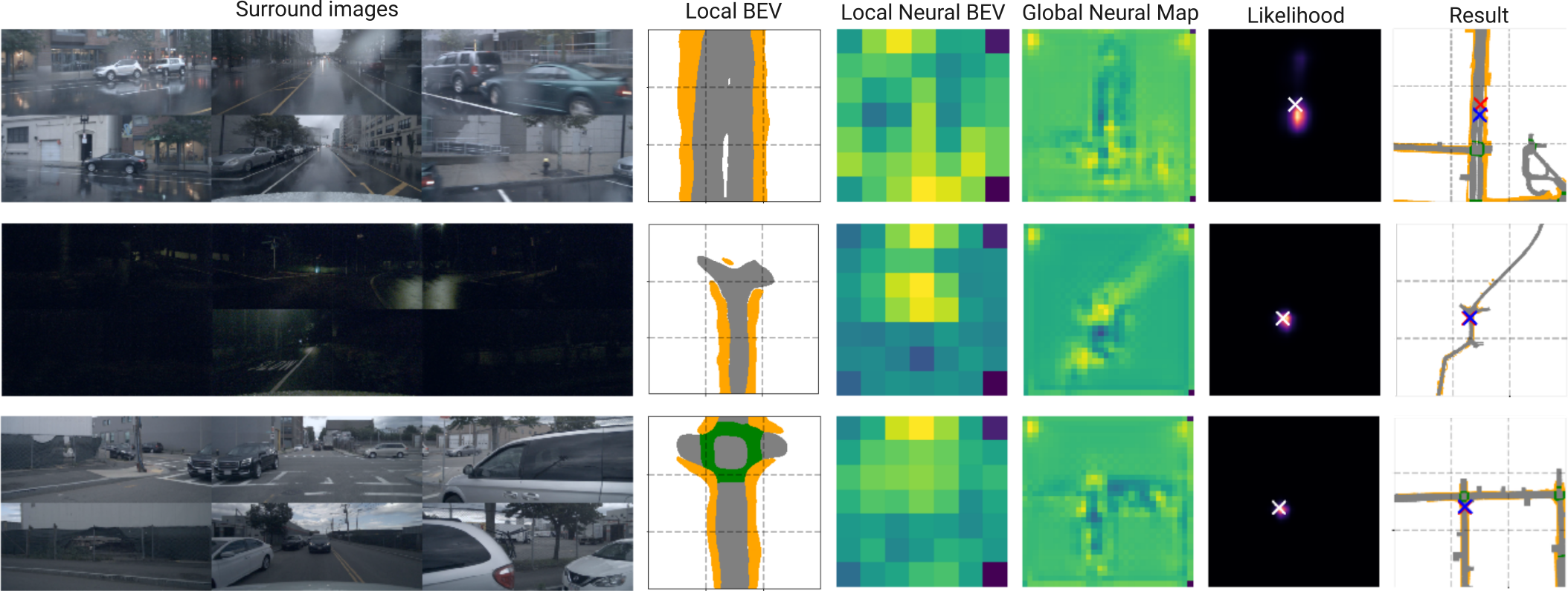}
\caption{Example of inputs and outputs of the localization pipeline, including surround images, local BEV, neural encoding of local BEV and map tile, likelihood of the prediction, and visualization of the result. In the visualization, the blue cross is the ground truth pose while the red cross is the prediction.}
\label{fig:localization-samples}
\end{figure*}

\begin{table*}[!ht]
  \centering
    \resizebox{0.95\textwidth}{!}{
    \begin{tabular}{@{}lllllllll@{}}
    \toprule    
    Type    & BEV ($m$) & map ($m^2$) & cm/px & Layers & R@$1m$ & R@$2m$ & R@$5m$ & R@$10m$ \\ \midrule
    OrienterNet    & 60   & 300$\times$300   & 75    & 3      & 17.29 & 38.22   & 57.21   & 65.11 \\
    OrienterNet    & 60   & 300$\times$300   & 30    & 3      & \textbf{31.23} & \textbf{46.90}   & 58.89   & 66.04 \\
    OrienterNet only drivable   & 60   & 300$\times$300   & 30    & 1      & \underline{24.08} & 38.42   & 54.63  & 62.20 \\
    CVT Neural Relocalization & 60 & 300$\times$300   & 30    & 3 & 7.691 & 23.61 & 58.44 & 77.03 \\
    U-BEV only drivable (Ours)   & 60     & 300$\times$300 & 30    & 1 & 13.69     & 37.67   & \underline{69.57}   & \underline{82.78} \\
    U-BEV Neural Relocalization (Ours)   & 60     & 300$\times$300 & 30    & 3 & 16.89     & \underline{41.60}   & \textbf{71.33}   & \textbf{83.46}
    \end{tabular}
    }
    \caption{Localization results in terms of Recall Accuracy at $1m,2m,5m,10m$.}
    \label{tab:largelocal}
\end{table*}

\subsection{Results}

\textbf{BEV:} Table~\ref{tab:bev} summarizes the BEV results of U-BEV and CVT. U-BEV outperforms CVT in all categories in terms of IoU. While the performance on drivable surfaces improves by 1.7 IoU, the improvement is much higher on Walkways and Crossing with an improvement of 2.8 and 2.3 respectively. Notably, U-BEV achieves this performance boost at a reduced computational complexity, requiring 5\% less GFLOPs during inference. Single-class inference consistently improves U-BEV by up to 2 IoU. These results indicate that the representational power of the proposed U-BEV is considerably larger than the Transformer-based CVT baseline. We attribute this to firstly, the multi-height encoding giving a boost of up to 1.7 IoU compared to the single-height case. Secondly, U-BEV retains full resolution through skip connections while CVT upsamples from a bottleneck of size $25\times 25$. Sample results in Fig.~\ref{fig:ubev-samples} demonstrate outputs from our system, including height predictions, and BEVs from U-BEV and CVT. Notably, U-BEV shows superior performance in drivable surface and sidewalk segmentation.

\newcommand{\da}{$\downarrow$}
\newcommand{\ua}{$\uparrow$}
\begin{table}[H]
  \centering
    \begin{tabular}{@{}lllll@{}}
    \toprule
    Model            & GFLOPs \da         & Drivable \ua  & Walkways \ua  & Crossings \ua \\ \midrule
    CVT              & 18.7               & 65.3          & 25.4          & 26.1          \\
    CVT mc & 18.7               & 62.0          & 29.7          & 27.4          \\
    U-BEV, w/o H & \textbf{17.16}     & 66.23         & 30.8          & \underline{28.0}\\
    U-BEV mc light & \underline{17.81} & 66.08         & 30.96         & 26.63          \\
    U-BEV mc, w/o H & \textbf{17.16} & 62.69 & 27.93    & 24.5           \\
    U-BEV mc & 21.46    & \underline{66.8} & \underline{32.0} & 27.7       \\
    U-BEV      & \underline{17.81}  & \textbf{67.0} & \textbf{32.5} & \textbf{29.7}  \\
    \end{tabular}%
    \caption{IoU performance on the \cite{goodsplit} split with $100\times100$m and $50$cm per pixel. mc refers to multi-class models, w/o H refers to models without height.}
    \label{tab:bev}

\end{table}

\textbf{Relocalization:}
Table~\ref{tab:largelocal} summarizes the relocalization results. OrienterNet performs best in the R@$1m$ and R@$2m$ bins which we attribute to the more sophisticated matching pipeline at full resolution. OrienterNet is good at picking up distinct features of the environment but suffers under semantically poor regions. U-BEV relocalization on the other side largely outperforms the baselines in the R@$5m$ and R@$10m$ bins which we attribute to its ability to reason about the shape of roads over the presence of distinct features. We argue that the good performance in the R@$5m$ and R@$10m$ bins is highly relevant for a lightweight relocalization system in diverse scenarios, including feature-poor regions. Task-specific local perception systems can bridge the remaining error to a desired accuracy as elaborated in Section~\ref{sec:intro}. CVT-based relocalization with the proposed template matcher has the lowest performance in the lower bins but improves on the R@$10m$ bin. U-BEV increases performance by 21.1\% R@$5m$ and 26.4\% R@$10m$ over the best baselines. We also perform experiments with only drivable surfaces in the map data. Here, U-BEV suffers only a minor performance decrease, while the baseline degrades especially in the R@$1m$ and R@$2m$ bins. We attribute this to the ability of U-BEV to rely more on the scene structure than on distinct semantic landmarks as the baseline.

\section{CONCLUSIONS}
This paper proposes a novel U-Net inspired BEV architecture ``U-BEV" that leverages intermediate Segmentation on multiple heights. This architecture outperforms contemporary transformer-based BEV architectures in Segmentation performance by $1.7$ to $2.8$ IoU. In addition, we propose a novel relocalization approach that leverages the proposed U-BEV for localization against neurally encoded SD map data together with Deep Template Matching. The relocalization extension largely outperforms related methods by over $26\%$ in Recall Accuracy at $10m$. Notably, the need for only a few classes in map data, especially the road surface, paves the way to relocalization in feature-poor environments.

\printbibliography

\end{document}